\documentclass[runningheads]{llncs}
\usepackage{graphicx}
\usepackage{amsmath}
\usepackage{amssymb}
\usepackage{fontawesome5}
\usepackage[utf8]{inputenc}
\usepackage{csvsimple}
\usepackage{enumitem}
\usepackage{float}
\usepackage{tabularx}

\usepackage{booktabs}

\usepackage{tikz}
\usetikzlibrary{snakes}
\usepackage{soul,color}

\newcommand{\nudgeSections}{0cm}
\newcommand{\nudgeSubsections}{0cm}

\begin{document}

\title{Winning at Any Cost - Infringing the Cartel Prohibition With Reinforcement Learning\thanks{The work presented in this paper has been conducted in the \emph{KarekoKI} project, which is funded by the \emph{Baden-Württemberg Stiftung} in the \emph{Responsible Artificial Intelligence} program.}}
\titlerunning{Winning at Any Cost}
\author{Michael Schlechtinger\inst{1}\orcidID{0000-0002-4181-3900} \and
Damaris Kosack\inst{2}\orcidID{0000-0002-7599-3233}\and
Heiko Paulheim\inst{1}\orcidID{0000-0003-4386-8195}\and
Thomas Fetzer\inst{2}\orcidID{0000-0002-5148-4610}}
\authorrunning{M. Schlechtinger et al.}
\institute{University of Mannheim, Chair of Data Science, 68159 Mannheim, Germany\and
University of Mannheim, Chair of Public Law, Regulatory Law and Tax Law, 68159 Mannheim, Germany}
\maketitle              % typeset the header of the contribution
\begin{abstract}Pricing decisions are increasingly made by AI. Thanks to their ability to train with live market data while making decisions on the fly, deep reinforcement learning algorithms are especially effective in taking such pricing decisions. In e-commerce scenarios, multiple reinforcement learning agents can set prices based on their competitor's prices. Therefore, research states that agents might end up in a state of collusion in the long run. To further analyze this issue, we build a scenario that is based on a modified version of a prisoner's dilemma where three agents play the game of rock paper scissors. Our results indicate that the action selection can be dissected into specific stages, establishing the possibility to develop collusion prevention systems that are able to recognize situations which might lead to a collusion between competitors. We furthermore provide evidence for a situation where agents are capable of performing a tacit cooperation strategy without being explicitly trained to do so.

\end{abstract}
\section{Introduction}
\vspace{\nudgeSections}
Dynamic reinforcement learning based pricing strategies supersede static ones in terms of average daily profits \cite{Kropp.2019}. As 27 percent of the respondents of a 2017 study by KPMG identified price or promotion as the factors that are most likely to influence their decision regarding which product or brand to buy online \cite{KPMG.}, it is to be expected that successful companies (such as Amazon \cite{Chen.2016}) base their decisions on these algorithms to learn from and react to their competitor’s pricing policies as well as to adjust to external factors, such as a transformation of demand or product innovations \cite{Ezrachi.2017}. Monitoring these AIs is getting increasingly complex as the market is distributed worldwide, the barriers of entry are minimal, and the amount of created pricing data grows quicker by the day. 

Primarily legal scholars have commented on the possibility of self-learning algorithms to quickly learn to achieve a price-setting collaboration especially within oligopolies (e.\,g., \cite{Ezrachi.2016,Ezrachi.2015,Ezrachi.2017}). With the power of modern hardware, AIs would be able to monitor the market in which they act, resulting in a rapidly arising tacit collusion. Researchers investigated the issue by creating game theory like scenarios with the intention of pushing the agents towards a Nash equilibrium (e.\,g., \cite{Ezrachi.2017,Waltman.2008}). In essence, it seems to be “incredibly easy, if not inevitable” to achieve “such a tacitly collusive, profit-maximizing equilibrium” \cite{Schwalbe.2018}. While collusion has been presumed to appear in enclosed multi agent reinforcement learning scenarios, scholars have neither studied how to spot the origin of collusion nor if competitors can apply tacit collusion by displacing the others.

In an effort to simplify the dynamic pricing data analysis, we aim to train a competitive multi agent reinforcement learning (MARL) game simulation. In this game, the agents play a three-player version of rock paper scissors (RPS). We aim to analyze the effect of the competitive RPS scenario on the agents’ learning performances and potential collaboration strategies. In specific, we aspire to analyze whether RL agents are capable of performing a tacit cooperation or communication strategy without being explicitly trained to do so.

% The remainder of this paper is structured as follows. Section 2 introduces the term of trustworthy AI in the context of reasonable machines and white box AIs. We furthermore present a foundation of anti-competitive agreements and possible overlaps between Multi Agent Reinforcement Learning, and related work in analyzing possible cooperation between reinforcement learning agents. In section 3, we establish a case study that builds on Deep Q-Network-Agents to present the training process of collusion. Section 4 portrays the outcomes of the different learning sessions. Finally, we discuss our findings, highlight the boundaries of our study, and conclude with promising avenues for future research.

\section{Related Work}
\vspace{\nudgeSections}

\subsection{Infringing the Cartel Prohibition}
\vspace{\nudgeSubsections}
In its most recent proposal for an Artificial Intelligence Act, the European Commission emphasises the importance of the safety and lawfulness of AI systems, of legal certainty with regard to AI, the governance and effective enforcement of existing law on fundamental rights and the installation of safety requirements \cite{EuropeanCommission.20210421}. In line with these goals, AI price policies must oblige to competition law just as prices that are set by humans. Both European and German competition law distinguish three possible conducts of infringing the cartel prohibition, see Article 101 (1) Treaty on the Functioning of the European Union ("TFEU")\footnote{Corresponding provision under German law: § 1 Act against Restraint of Competition; corresponding provision under US law Section 1 Sherman Antitrust Act of 1890.}: (a) \emph{agreements between undertakings}, (b) \emph{decisions by associations of undertakings}, and (c) \emph{concerted practices}. Independent undertakings shall independently decide over their market behavior and must not coordinate it with their competitors (“requirement of independence”). This requirement does strictly preclude any direct or indirect contact by which an undertaking may influence the conduct on the market of its actual or potential competitors or disclose to them its decisions or intentions concerning its own conduct on the market where the object or effect of such contact is to create conditions of competition which do not correspond to the normal conditions of the market \cite{EuropeanCourtofJustice.}. 

The independently chosen intelligent adaption of an undertaking's market behavior to the observed market behavior of its competitors (generally) is permitted. Drawing a clear line between the adaption of an observed market behavior and a conduct through which competition knowingly is replaced by a practical cooperation and therefore constitutes a concerted practice within the meaning of Article 101 (1) TFEU\footnote{For US law see \cite{DOJandFTC.2123June2017,Gulati.}} is often difficult and sometimes even impossible. Especially on transparent markets with few market participants, the market outcome of collusion can often hardly be traced back to be (or not to be) the product of a concerted practice (cf. petrol station market). Although collusion as a market outcome can be detrimental to consumers, innovation and economic growth and is therefore undesirable from a welfare economic point of view, the difficulty from a dogmatic perspective is that legal responsibility cannot be attached to a market outcome as such \cite{Weche.2020}.

Our goal is to disclose whether a certain sequence of actions or a specific pattern can be identified as a situation in which the uncertainty about the competitor's next moves is replaced by a practical cooperation. It is conceivable that such accurate determination might not be possible due to the increased market transparency achieved by the self-learning algorithms: their ability to quickly process large amounts of competition-relevant data and to react to price movements in an almost unlimited frequency might lead to such a high degree of transparency on a market that makes it impossible to determine from its outcome whether or not the result of collusion is due to intelligent market observation and parallel behavior or a concerted practice. 

\subsection{Multi Agent Reinforcement Learning}
\vspace{\nudgeSubsections}
A tacit collaboration between some reinforcement learning agents can only occur in certain situations. The agents have to interact within a multi agent reinforcement learning (MARL) environment, where competing agents and prices are recognized as a part of such \cite{Charpentier.2020}. Due to that, the environment is usually subjective for every agent, resulting in a differing learning performance and a diverse landscape of achieved competencies. It is unclear whether one of these competencies might arise in the skill to communicate with specific other agents to adjust their pricing policies accordingly; resulting in a higher producer’s pension and a displacement of a competitor.

%Researchers have investigated circumstances which can be juxtaposed with collusion between pricing agents. As such, Schwind \cite{Schwind.2007} successfully modeled a bidding process as a MARL problem. Competitors’ bids as well as other auction information were represented in every agent’s subjective environment, while its action space is confined to the bid price. However, just like in other multi agent simulations, the scalability regarding the number of agents is limited. Dütting et al. \cite{Dutting.12.06.2017} suggested to take advantage of deep neural networks to handle the state value approximation and thus increase scalability. A recent study by Zheng et al. \cite{Zheng.28.04.2020} set new foundations for future research on MARL economy simulations with multiple actors in different roles. Supported by a two-level (state and taxpayer) deep reinforcement learning approach, the agents successfully learn and adapt to dynamic tax policies. However, the authors did not control for or induce communication or collaboration between agents.

Researchers have investigated circumstances which can be juxtaposed with collusion between pricing agents, such as bidding processes \cite{Schwind.2007,Dutting.12.06.2017} or economy simulations \cite{Zheng.28.04.2020}. However, the authors did not control for or induce communication or collaboration. To combat this shortcoming, scholars within the economics realm created oligopolistic models (particularly Cournot oligopolies) to show collusion between agents. A Cournot oligopoly is characterized by an imperfect competition, where firms individually have some price-setting ability but are constrained by rivals \cite{AugustinA.Cournot.1836}. Izquierdo and Izquierdo \cite{IZQUIERDO.2015} show that simple iterative procedures, such as the win-continue, lose-reverse (WCLR) rule are able to achieve collusive outcomes. However, the results are not robust in terms of minor, independent perturbations in the firms’ cost or profit functions. Similar results were achieved with basic Q-learning \cite{Waltman.2008}. As a case in point, using a price-setting duopoly model with fixed production, in which two firms follow a Q-learning algorithm, Tesauro and Kephart \cite{Tesauro.2002} observed convergence to prices higher than the competitive level. Major takeaways from these studies are, that cooperation is more likely to occur in simplified, static environments with a homogeneous good and that communication is vital to achieve collusive outcomes, particularly when more than two firms operate in a market. Such results suggest that the ability to communicate could also be pivotal for algorithmic collusion to occur \cite{Schwalbe.2018}.

\section{Methodology}
\vspace{\nudgeSections}
\subsection{Problem Definition}
\vspace{\nudgeSubsections}
Oroojlooy and Hajinezhad \cite{OroojlooyJadid.11.08.2019} recommend to model a MARL problem based on (i) centralized or decentralized control, (ii) fully or partially observable environment and (iii) cooperative or competitive environment. Our case demands for a decentralized control, with a partially to fully observable environment, so that every agent is able to make its own decisions based on the information given by the environment. Lastly, we apply a cooperative inside of a competitive environment, so that agents are able to team up against other agents.

\subsection{Approach}
\vspace{\nudgeSubsections}

\begin{table}[t]
\centering
\caption{Three player RPS combinatorics.}\label{tabRPS}
\newcolumntype{C}{>{\centering\arraybackslash}X}
\def\arraystretch{1.05}
\begin{tabularx}{0.65\textwidth}{ | l | l | l || C | C | C | }
\hline
{\bfseries Agent 1} & {\bfseries Agent 2} & {\bfseries Agent 3} & {\bfseries \(r_1\)} & {\bfseries \(r_2\)} & {\bfseries \(r_3\)}\\
\hline
Rock & Paper & Scissors         & 0 & 0 & 0\\
Rock & Rock & Rock              & 0 & 0 & 0\\
Scissors & Scissors & Scissors  & 0 & 0 & 0\\
Paper & Paper & Paper           & 0 & 0 & 0\\
Scissors & Rock & Rock          & -1 & 0.5 & 0.5\\
Rock & Paper & Paper            & -1 & 0.5 & 0.5\\
Paper & Scissors & Scissors     & -1 & 0.5 & 0.5\\
Paper & Rock & Rock             & 2 & -1 & -1\\
Rock & Scissors & Scissors      & 2 & -1 & -1\\
Scissors & Paper & Paper        & 2 & -1 & -1\\
... & ... & ...        & ... & ... & ...\\
\hline
\hline
\multicolumn{3}{|l||}{Expected Reward \(r\)} &  0 & 0 & 0\\
\hline
\end{tabularx}
\end{table}

With the intention of simplifying a realistic economy simulation, we choose to build a MARL-game based on a three player version of RPS. Every agent \(i = \{1,\,. . . ,\,3\}\) represents a player with a set of legal game actions \(A = \{1,\, . . . ,\,3\}\) comprising the moves of rock, paper and scissors. The agents interact with a stochastic environment \(E\) which solely contains the chosen actions of every agent of the current time step \(t\). Hence, a state at \(t\) can be described as \(s_t = \{a_1',\,. . .,\,a_i'\}\). Following a collective action, every agent receives a reward out of \(R = \{\text{-}1,\,0,\,0.5,\,2\}\) mapped to the possible game outcomes presented in table \ref{tabRPS}, resulting in a direct association between input and output. This formalism gives rise to a finite Markov decision process (MDP) in which every \(t\) relates to a distinct state, encouraging an application of standard reinforcement learning methods for MDPs. The goal of the agent is to interact with \(E\) by selecting actions in a way that maximises future rewards. As the agents receive a reward at the end of every timestep, we will not apply any discount to future rewards. We define the optimal action-value function \(Q^\ast(s, a)\) as the maximum expected return achievable by following any strategy, after seeing some sequence \(s\) and then taking some action \(a\), \(Q^\ast (s, a) = \max_{\pi} \mathbb{E} [R_{t}|s_t = s, a_t = a, \pi]\), where $\pi$ is a policy that maps sequences to actions (or distributions over actions). In an attempt to induce strategic behavior, resulting in a tacit communication within this competitive MARL environment, we utilize a Deep Q-Network (DQN) \cite{Mnih.19.12.2013} with an experience replay and a target network \cite{L.Lin.1992}. After performing experience replay, the agent selects and executes an action according to an $\epsilon$-greedy policy. The agents select the action \(a^t\) that maximizes the expected value of \(r + Q^\ast (s', a' )\), updating the Q-values by:
\begin{equation}
 Q^\ast (s, a) = \mathbb{E}_{s'\sim \varepsilon} [r + \underset{a'}{\max} \, Q^\ast (s' , a' ) | s, a]
\end{equation}
Our main argument for the selection of this specific scenario is the controlled, unambiguous reward allocation in combination with the restricted moveset of the agents. Thus, every step \(t\) develops into a zero-sum game (as shown in Table \ref{tabRPS}). On the one hand, we create an environment, where no agent can earn a profit, if it does not communicate with another agent. On the other hand, we counteract the poor learning performance of MARL \cite{Allen.2009} (due to the defined equilibrium/ local optimum) as well as increase the comprehensibility of the neural network's predictions. We expect the agents to converge to a collusive state after several episodes, as described by economics and law scholars (e.\,g., \cite{Ezrachi.2017,Waltman.2008}).

We also attempt to induce a displacement of one agent due to the actions selected by the other two agents. In our use case, they need to learn a specific policy which would force two colluding agents to not repeat their allied agents' actions. While this would not necessarily result in a better short term step reward for these agents, it would however eliminate the ability to achieve a "big win" (e.\,g., playing Paper if the opponents play Rock and Rock) for the third, competing agent. Generally speaking, if two agents avoid choosing the same action, the expected reward for the third player is negative. We aim to simulate this circumstance in diverging settings. 

In \textit{mode 1}, collusion is induced by explicit communication as suggested by Schwalbe \cite{Schwalbe.2018}. More specifically, we designate two 'cheating' agents \(i_{c} \subset i\) and a 'fair' agent \(i_{f} \in i,\, i_{f} \not\in i_{c}\) ahead of a training session. Before its turn, one of the cheating agents transmits his picked action to the other cheating agent. The message will be enclosed to input to the receiver's DQN.\footnote{It is important that in the eyes of the receiving agent, this is just a variable with the values of \(A\) which does \emph{not} have the specific semantics of \emph{this is the other agent's next move}.} In \textit{mode 2}, instead of making the players communicate explicitly, we provoke tacit communication by adjusting the reward of the cheating agents \(r^{t}_{i_c}\) to \(r^{t}_{i_c} = -r^{t}_{f}\). In other words, they will try to maximize their \emph{joint} instead of their \emph{individual} reward, which is equivalent to minimizing $i_f$'s reward. We additionally \textit{denoise} the rewards; hence, \(i_{c}\) will receive 1 for a loss or a tie with \(i_{f}\) and -1 for a win of \(i_{f}\). To further stress this issue, we perform control-runs, where \(i_{f}\) is replaced with an agent that plays random actions (which is the best strategy in a competitive 3-player version of RPS). 

\subsection{Implementation}
\vspace{\nudgeSections}

\begin{figure}[t]
  \centering
    \includegraphics[trim=30 23 15 22,clip,width=\textwidth]{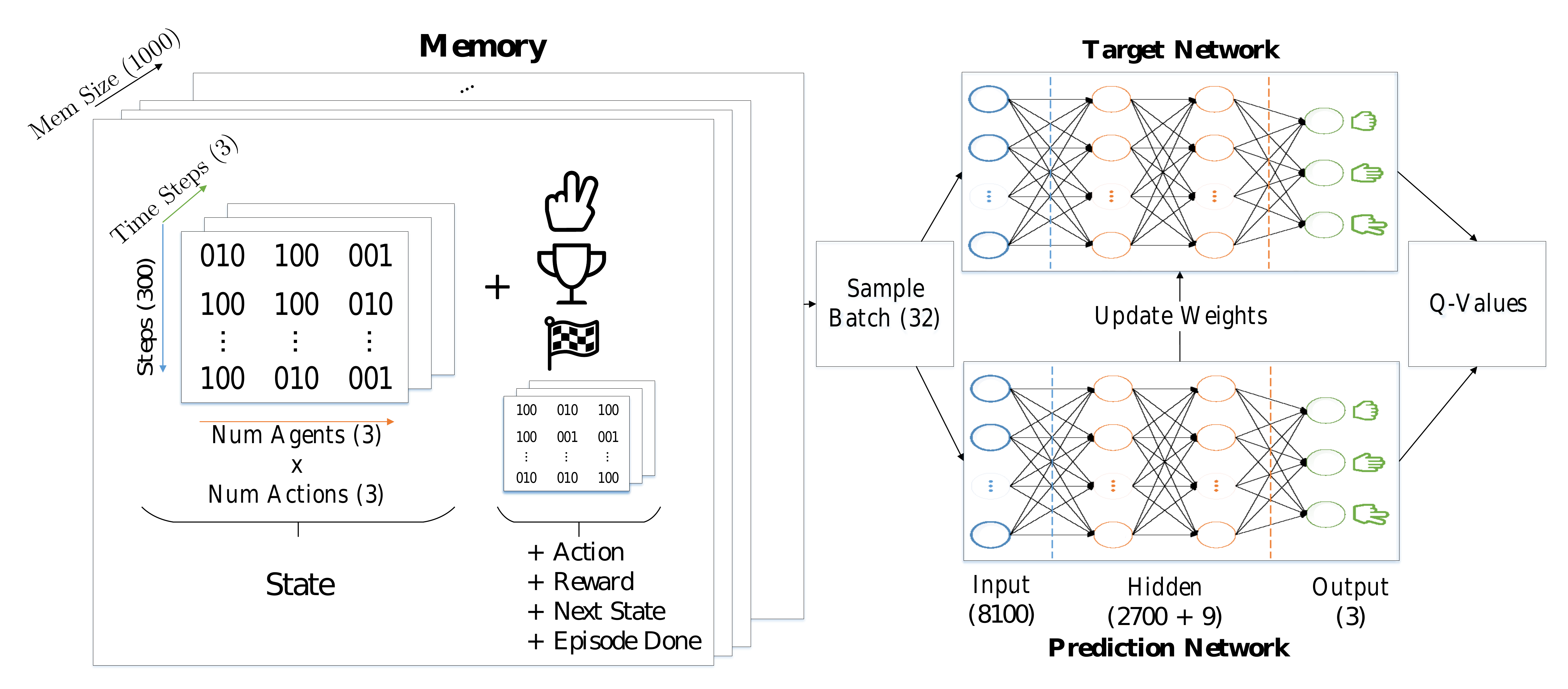}
  \caption{DQN Architecture} \label{dqnArch}
\end{figure}

The main weakness of RPS in a real world scenario is the unpredictability of an opponent's move. The best player would just play random, however since playing this game is psychologically based on personal human short-term memory behavior, there is a tendency to follow specific patterns, like not repeating moves or trying to play unpredictably \cite{Ali.2000}. In an artificial MARL-problem, we can model that by not only reacting to an opponent's last move, but learning from a history of its last moves. After testing, we chose to apply a history size of 100 games to accommodate for a stable learning process. Regarding the experience replay, we chose to use the last 3 timesteps as an input for the neural net. The network is made up of four dense layers (input, two hidden layers, output), whose main task is to compress the given information and provide the chosen action. For that matter, we design a DQN with an input layer comprising 8100 neurons (300 steps \textasteriskcentered{} 3 one-hot encoded actions \textasteriskcentered{} 3 players \textasteriskcentered{} 3 time steps), two hidden layers with 2700 and 9 neurons and a dense output with 3 neurons to choose either rock, paper or scissors (cf. figure \ref{dqnArch}). The neurons required for the number of players will increase by 1 for the cheating player to accommodate for the action received by the messaging agent. We use TensorFlow 2 to build and train the DQNs; the code can be found on Github\footnote{https://gitfront.io/r/user-7017325/1eb2ef3332343def1c7f67d5fce5953f1e003681/\hspace{3pt}AiCollusionDQN/}.

\section{Results}
\vspace{\nudgeSections}
To counter inconsistent MARL outcomes, we chose to train the agents for 10 runs with 100 episodes each (300 steps per episode), comprising three different learning rates (0.001, 0.005, 0.01), resulting in 30 runs with 900.000 games of RPS per scenario. We picked learning rates that are fairly small to counteract quickly developing local optima, causing repetitions of the same action, due to the straightforward connection between action and reward. For every scenario with \(i_{c}\) involved, we also performed another 15 runs (5 per learning rate) where \(i_{f}\) is replaced with an agent that randomly picks actions in order to further stress the issue by simulating a perfect RPS policy.

\subsection{Collusion between all agents}
\vspace{\nudgeSubsections}
\begin{figure}[t]
  \centering
%   \begin{minipage}[b]{0.49\textwidth}
%     \includegraphics[width=\textwidth]{figures/CollusionAll/lr001.pdf}
%     %\caption{Learning Rate 0.001}
%   \end{minipage}
%   \hfill
%   \begin{minipage}[b]{0.49\textwidth}
%     \includegraphics[width=\textwidth]{figures/CollusionAll/lr005.pdf}
%     %\caption{Learning Rate 0.005}
%   \end{minipage}
%   \begin{minipage}[b]{0.49\textwidth}
    % \includegraphics[width=\textwidth]{figures/CollusionAll/lr01.pdf}
    %\caption{Learning Rate 0.001}
%   \end{minipage}
%   \hfill
%   \begin{minipage}[b]{0.49\textwidth}
    \includegraphics[width=0.95\textwidth]{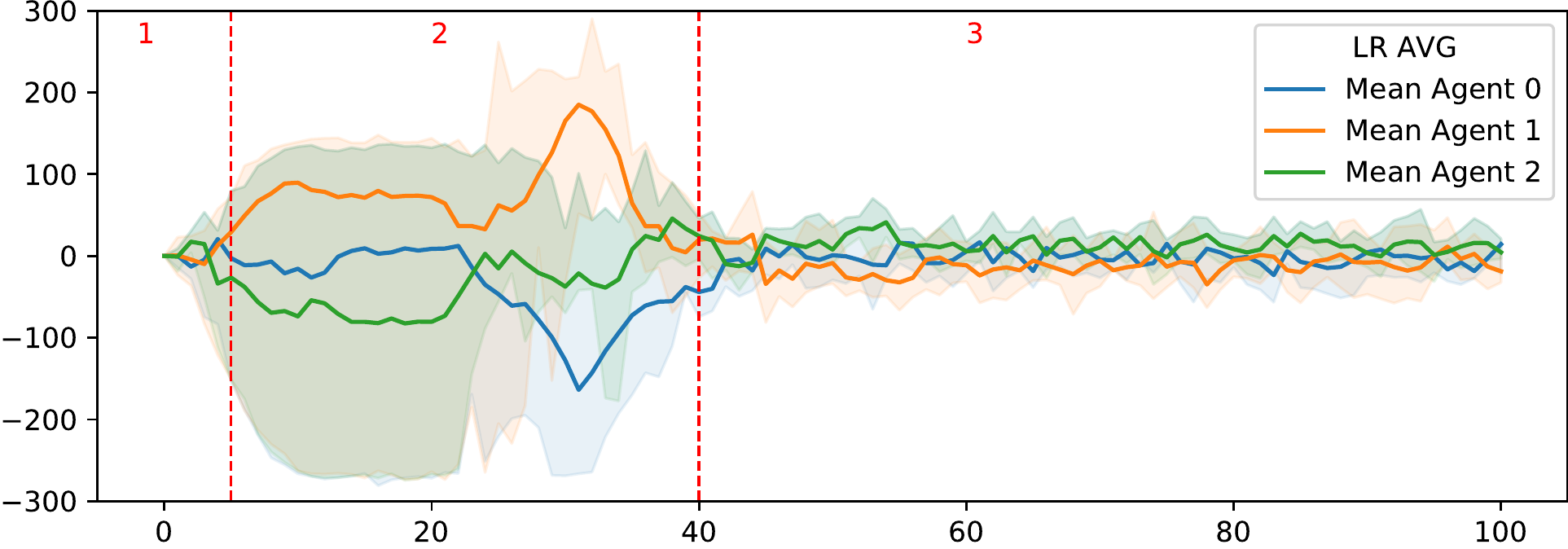}
    %\caption{Learning Rate 0.005}
%   \end{minipage}
  \caption{Episode reward distribution within the different learning rate scenarios.} \label{rewardsLearningRate}
\end{figure}

\begin{table}[t]
\newcommand{\lowerEnd}{6}
\newcommand{\upperStart}{292}
\centering
\caption{Action samples from two different runs, divided in stages 1, 2, 3a and 3b}\label{tabStages}

\hspace{0.1cm}

\csvreader[head to column names, tabular=*{4}{|c}|,
table head=\hline step & A0 & A1 & A2\\\hline,
late after line=\\\hline,
%>> list only matriculation numbers greater than 20000 <<
filter expr=
    test{\ifnumless{\csvcoli}{\lowerEnd}}
    or test{\ifnumgreater{\csvcoli}{\upperStart}}, %\or \ifnumgreater{\csvcoli}{275}
]{./stageData/FaStage1.csv}{}{%
\csvcoli & \csvcolii & \csvcoliii & \csvcoliv}%
\hspace{0.7cm}
%
\csvreader[head to column names, tabular=*{4}{|c}|,
table head=\hline step & A0 & A1 & A2\\\hline,
late after line=\\\hline,
%>> list only matriculation numbers greater than 20000 <<
filter expr=
    test{\ifnumless{\csvcoli}{\lowerEnd}}
    or test{\ifnumgreater{\csvcoli}{\upperStart}}, %\or \ifnumgreater{\csvcoli}{275}
]{./stageData/FaStage2.csv}{}{%
\csvcoli & \csvcolii & \csvcoliii & \csvcoliv}%
\hspace{0.7cm}
%
\csvreader[head to column names, tabular=*{4}{|c}|,
table head=\hline step & A0 & A1 & A2\\\hline,
late after line=\\\hline,
%>> list only matriculation numbers greater than 20000 <<
filter expr=
    test{\ifnumless{\csvcoli}{\lowerEnd}}
    or test{\ifnumgreater{\csvcoli}{\upperStart}}, %\or \ifnumgreater{\csvcoli}{275}
]{./stageData/FaStage3a.csv}{}{%
\csvcoli & \csvcolii & \csvcoliii & \csvcoliv}%
\hspace{0.1cm}
%
\csvreader[head to column names, tabular=*{4}{|c}|,
table head=\hline step & A0 & A1 & A2\\\hline,
late after line=\\\hline,
%>> list only matriculation numbers greater than 20000 <<
filter expr=
    test{\ifnumless{\csvcoli}{\lowerEnd}}
    or test{\ifnumgreater{\csvcoli}{\upperStart}}, %\or \ifnumgreater{\csvcoli}{275}
]{./stageData/FaStage3b.csv}{}{%
\csvcoli & \csvcolii & \csvcoliii & \csvcoliv}%

% \csvreader[tabular=|c|c|c|c|,
%     table head=\hline step & A0 & A1 & A2 \\\hline,
%     late after line=\\\hline,
%     filter={\value{csvrow}<\noOfLines},
%     ]%
% {./stageData/FaStage3b.csv}{0=\0, 1=\1, 2=\2, 3=\3}%
% {\thecsvrow & \1 & \2 & \3}%

\vspace{0.25cm}%increase white space after table 

% Timeline
\begin{tikzpicture}[snake=zigzag, line before snake = 3mm, line after snake = 3mm]

\hspace{-0.35cm}
\newcommand{\spaceBelow}{5pt}

    % draw horizontal line   
    \draw (0,0) -- (1,0);
    \draw[snake] (1,0) -- (2,0);
    \draw (2,0) -- (4,0);
    \draw[snake] (4,0) -- (5,0);
    \draw (5,0) -- (8,0);
    \draw[snake] (8,0) -- (10,0);
    \draw (10,0) -- (11.5,0);

    % draw vertical lines
    \foreach \x in {0,3,6,11.5}
      \draw (\x cm,3pt) -- (\x cm,-3pt);

    % draw nodes
    \draw (0,0) node[below=\spaceBelow] {$ Episode\,0 $} node[above=3pt] {$ $};
    \draw (1.5,0) node[below=\spaceBelow] {$  $} node[above=3pt] {$ Stage\,1 $};
    \draw (2,0) node[below=\spaceBelow] {$ $} node[above=3pt] {$  $};
    \draw (3,0) node[below=\spaceBelow] {$ Episode\,5 $} node[above=3pt] {$  $};
    \draw (4,0) node[below=\spaceBelow] {$  $} node[above=3pt] {$ $};
    \draw (4.5,0) node[below=\spaceBelow] {$  $} node[above=3pt] {$ Stage\,2 $};
    \draw (6,0) node[below=\spaceBelow] {$ Episode\,40 $} node[above=3pt] {$  $};
    \draw (7,0) node[below=\spaceBelow] {$  $} node[above=3pt] {$  $};
    \draw (8,0) node[below=\spaceBelow] {$  $} node[above=3pt] {$ $};
    \draw (9,0) node[below=\spaceBelow] {$  $} node[above=3pt] {$ Stage\,3a)\And Stage 3b) $};
    \draw (10,0) node[below=\spaceBelow] {$  $} node[above=3pt] {$  $};
    \draw (11,0) node[below=\spaceBelow] {$ Episode\,100 $} node[above=3pt] {$  $};
    \draw (11.5,0) node[below=\spaceBelow] {$ $} node[above=3pt] {$  $};
\end{tikzpicture}
% \hspace{1cm}
\vspace{-0.6cm}%reduce too much white space after table 

\end{table}

In our series of simulations, we were able to achieve collusive results within every of the chosen learning rate scenarios (cf. figure \ref{rewardsLearningRate}). When averaged, the different action sequences can be visually divided into three learning stages. In \textbf{\textit{stage 1}}, the agents basically acted random, due to the epsilon-greedy algorithm. After approximately 5 episodes (\textbf{\textit{stage 2}}), one of the agents achieved a better outcome due to a lucky action selection. The agents stuck to their learned strategy while randomly delving into different policies. Upon further examination, we discovered that the strategies usually involve a single action which will be repeated in the next turns, even if this might not be the best action. This sub-optimal behavior stems from the first few episodes being mostly played randomly due to the epsilon-greedy strategy. Thus, the agents were only able to learn from a short history, which taught them to repeat the most successful action, rather than a certain sequence of actions. \textbf{\textit{Stage 3}} establishes a collusive sequence of actions from episode 40 onwards with two different scenarios (\textbf{3a} and \textbf{3b}).

As presented in table \ref{tabStages}, the agents try to avoid a negative reward over a long-term, resulting in an average episode profit of zero. However, stages 3a and 3b differ significantly in their way of achieving this. In 3a, one of the players repeated one action (e.\,g., scissors) and occasionally deviated from that due to the epsilon-greedy strategy, while the others predominantly alternate between two moves that change over time. In stage 3b, the agents played seemingly random. However if examined more closely, specific alternation patterns occured. A specific pattern can be identified, when observing the actions of agent 1 in figure \ref{tabStages}. The player oscillated between choosing rock and scissors in the first moves and transitions to scissors and paper towards the end of the episode. The remainder of the agents follow more elaborate patterns, however specific repetitions can be discovered by scanning the history.

\subsection{Collusion between two agents}
\vspace{\nudgeSubsections}

\begin{figure}[t]
  \centering
  \begin{minipage}[b]{0.49\textwidth}
    \includegraphics[width=\textwidth]{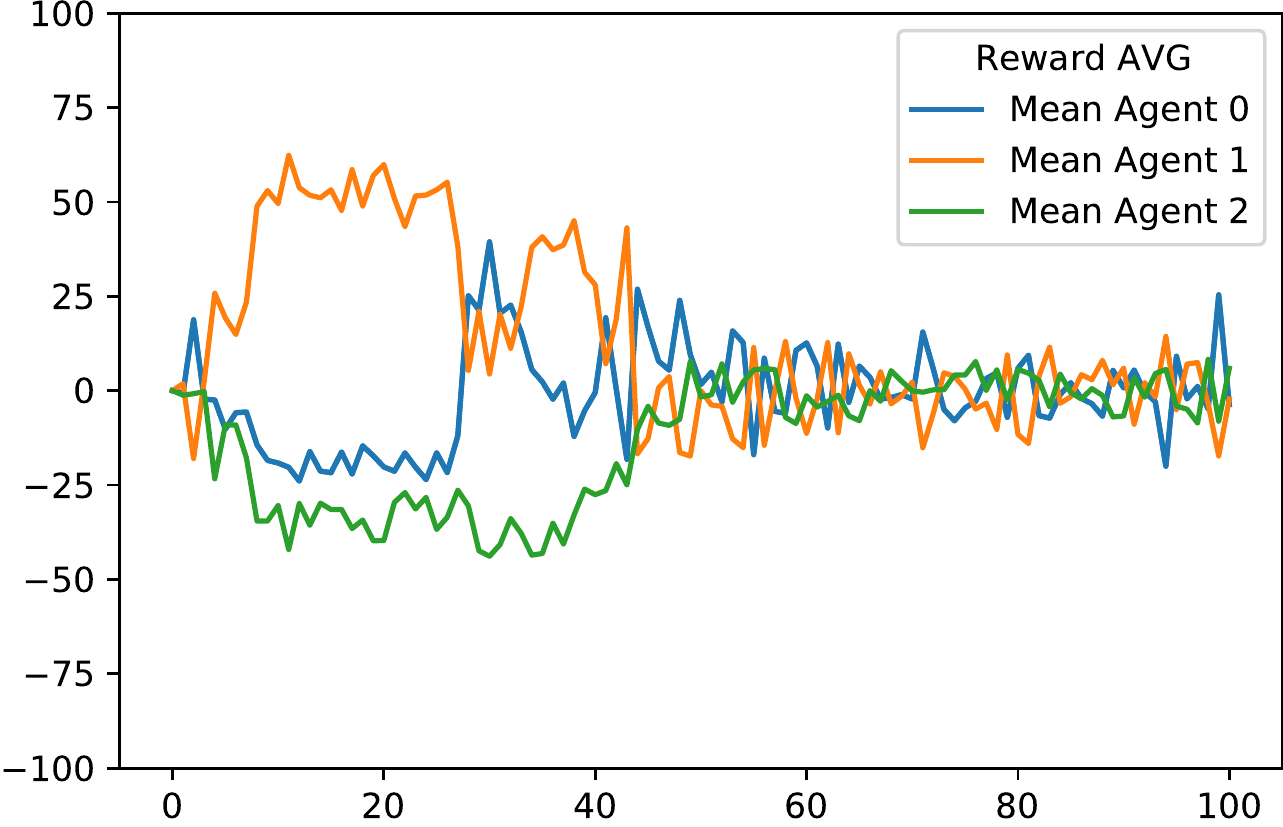}
    %\caption{Learning Rate 0.005}
  \end{minipage}
  \begin{minipage}[b]{0.49\textwidth}
    \includegraphics[width=\textwidth]{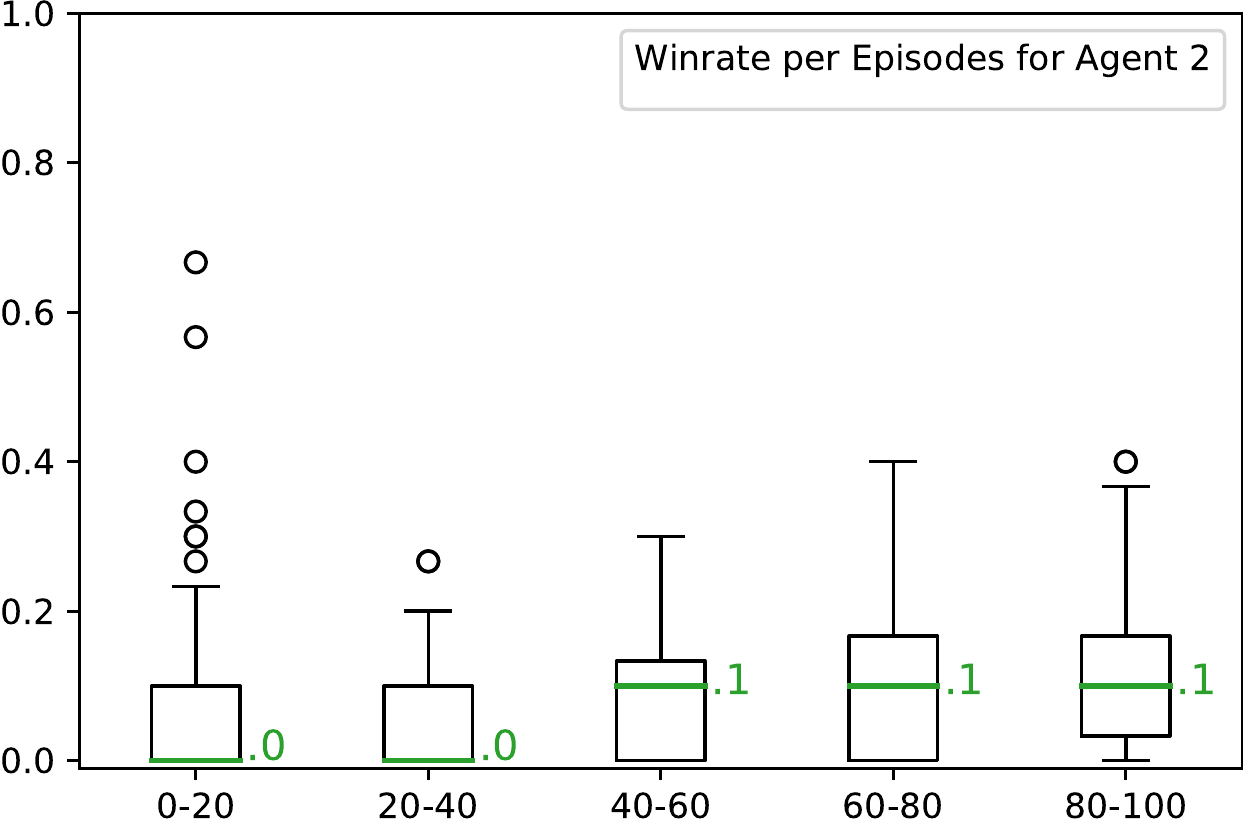}
    %\caption{Learning Rate 0.005}
  \end{minipage}
  \begin{minipage}[b]{0.49\textwidth}
    \includegraphics[width=\textwidth]{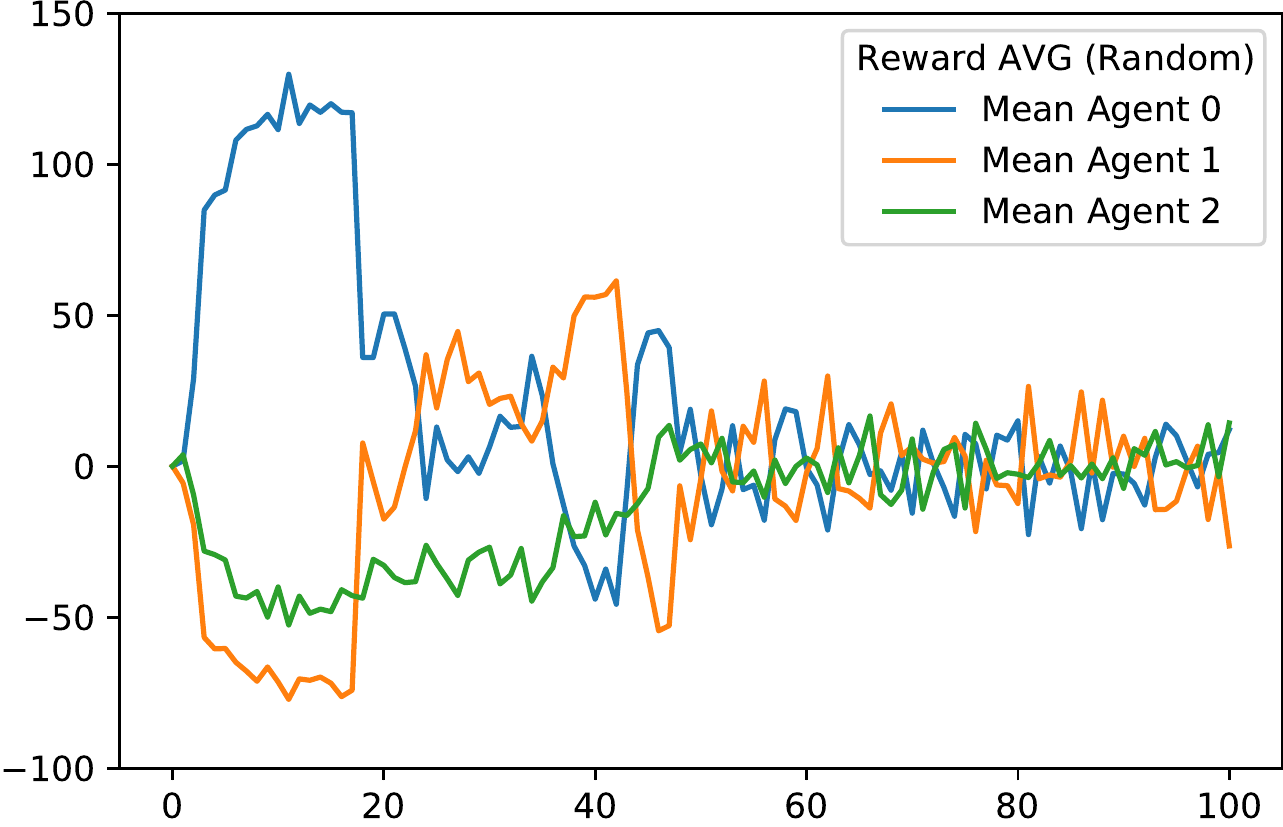}
    %\caption{Learning Rate 0.005}
  \end{minipage}
  \begin{minipage}[b]{0.49\textwidth}
    \includegraphics[width=\textwidth]{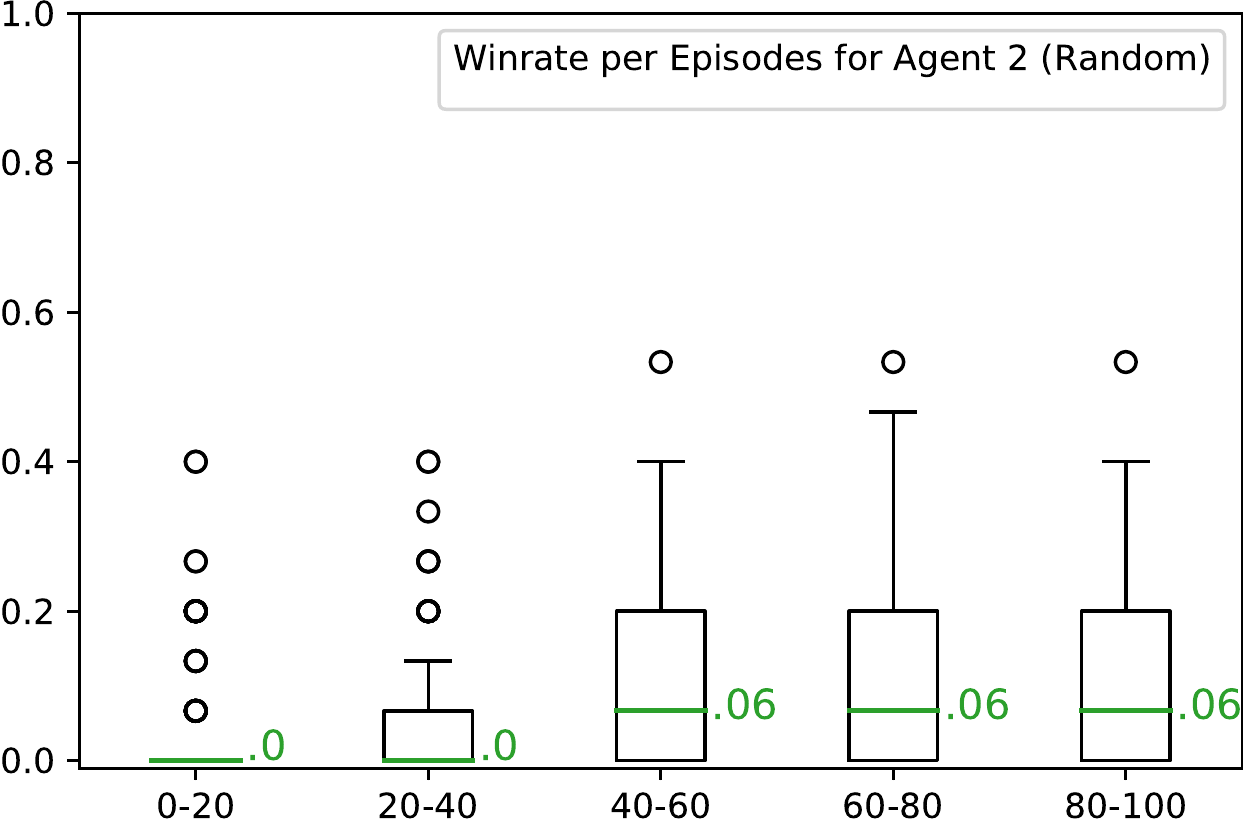}
    %\caption{Learning Rate 0.005}
  \end{minipage}
  \caption{Episode reward distribution within \textit{mode 1} including control runs where \(i_{f}\) is choosing random actions (lower half).} \label{collusionTwoAgentsExplicit}
\end{figure}

\subsubsection{Mode 1: Explicit Communication}
We successfully trained a displacing collusion policy with the help of explicit communication between the cheating agents \(i_{c}\). The results represented in figure \ref{collusionTwoAgentsExplicit} indicate that the agents were able to learn the suggested policy of not repeating their collaborator's action after a few episodes. After about 5 episodes, \(i_{c}\) achieve a higher reward on average. Thus, for the next 30 Episodes \(i_{f}\) is only rarely able to achieve a "big win". However, just like when colluding with all agents, after approximately 45 episodes, they tend to converge to an average game reward of 0. While it would be feasible to prolong this behavior by including variable learning rates or reducing the target update frequency during later episodes, we chose to encourage a long-term formation of a zero-centered equilibrium. Our reasoning behind this is the comparison to a real-world oligopoly, where two  market participants could only uphold a displacement by reducing the price up to a certain point, before damaging themselves.

In order to further stress the issue, we chose to replace \(i_{f}\) with an agent that chooses actions randomly. While \(i_{c}\) were able to successfully learn a displacement strategy in every training session, the results within the first 40 episodes were less significant than when \(i_{f}\) acted on behalf of the DQN. Nevertheless, we were able to observe slightly better results in the later stages, due to the added randomness.
%To discern the between the outcome of the game and the training progression, we differentiated between the  "actual" (denoised and negated) and the "game reward"

\begin{figure}[t]
  \centering
  \begin{minipage}[b]{0.49\textwidth}
    \includegraphics[width=\textwidth]{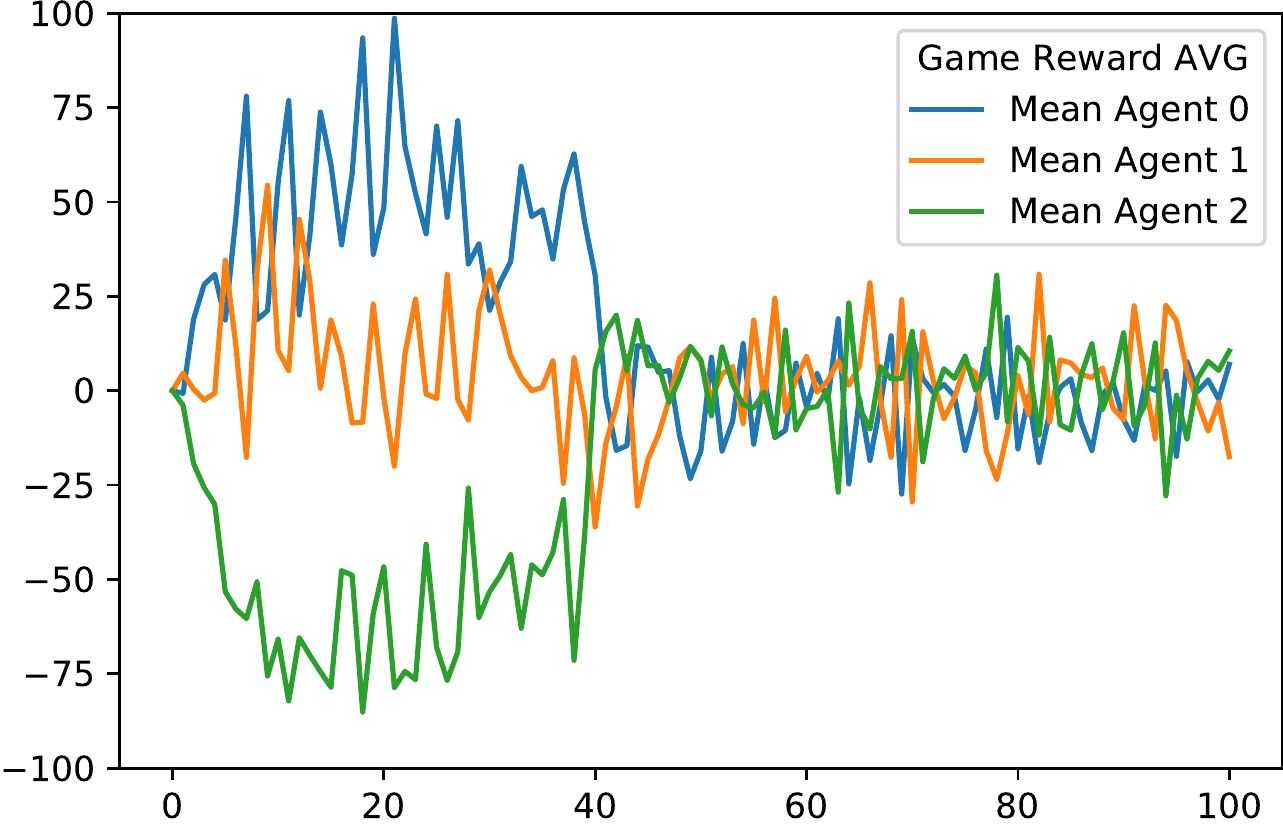}
    %\caption{Learning Rate 0.005}
  \end{minipage}
  \begin{minipage}[b]{0.49\textwidth}
    \includegraphics[width=\textwidth]{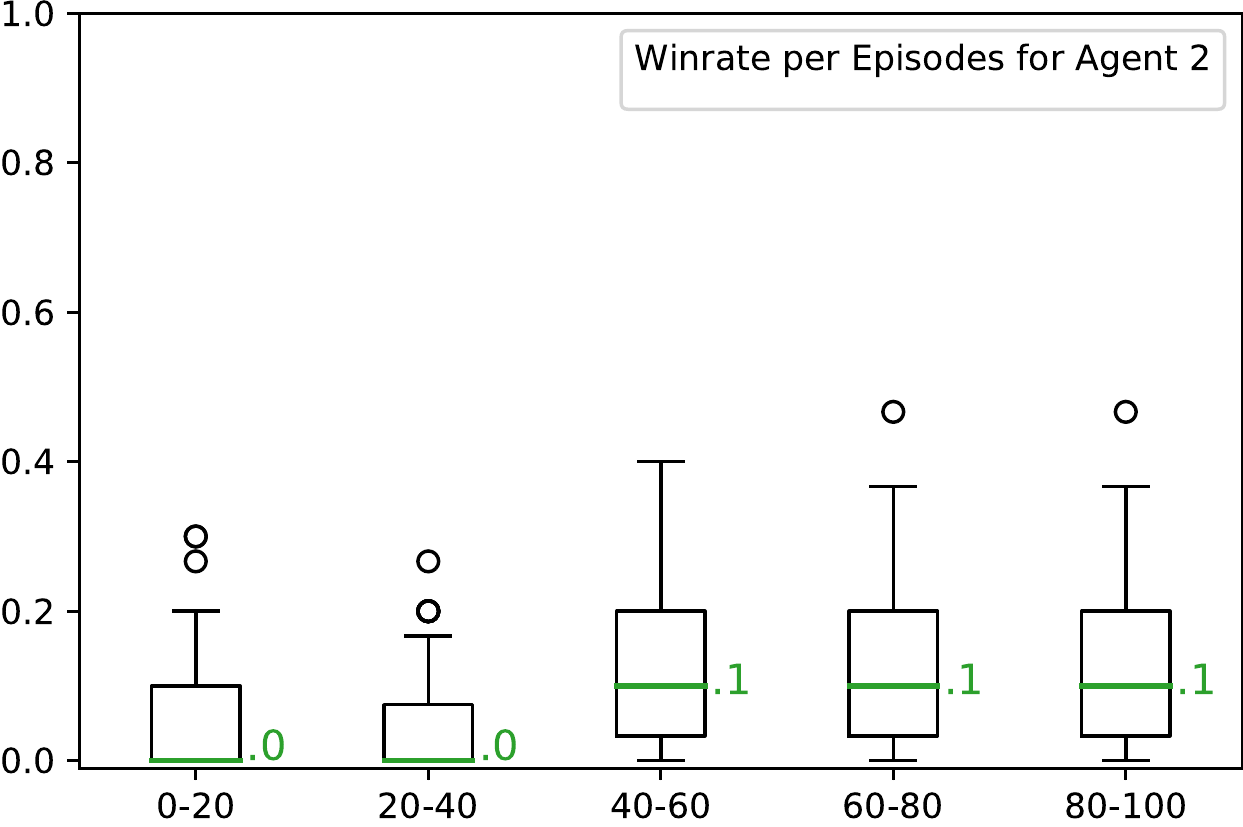}
    %\caption{Learning Rate 0.005}
  \end{minipage}
  \begin{minipage}[b]{0.49\textwidth}
    \includegraphics[width=\textwidth]{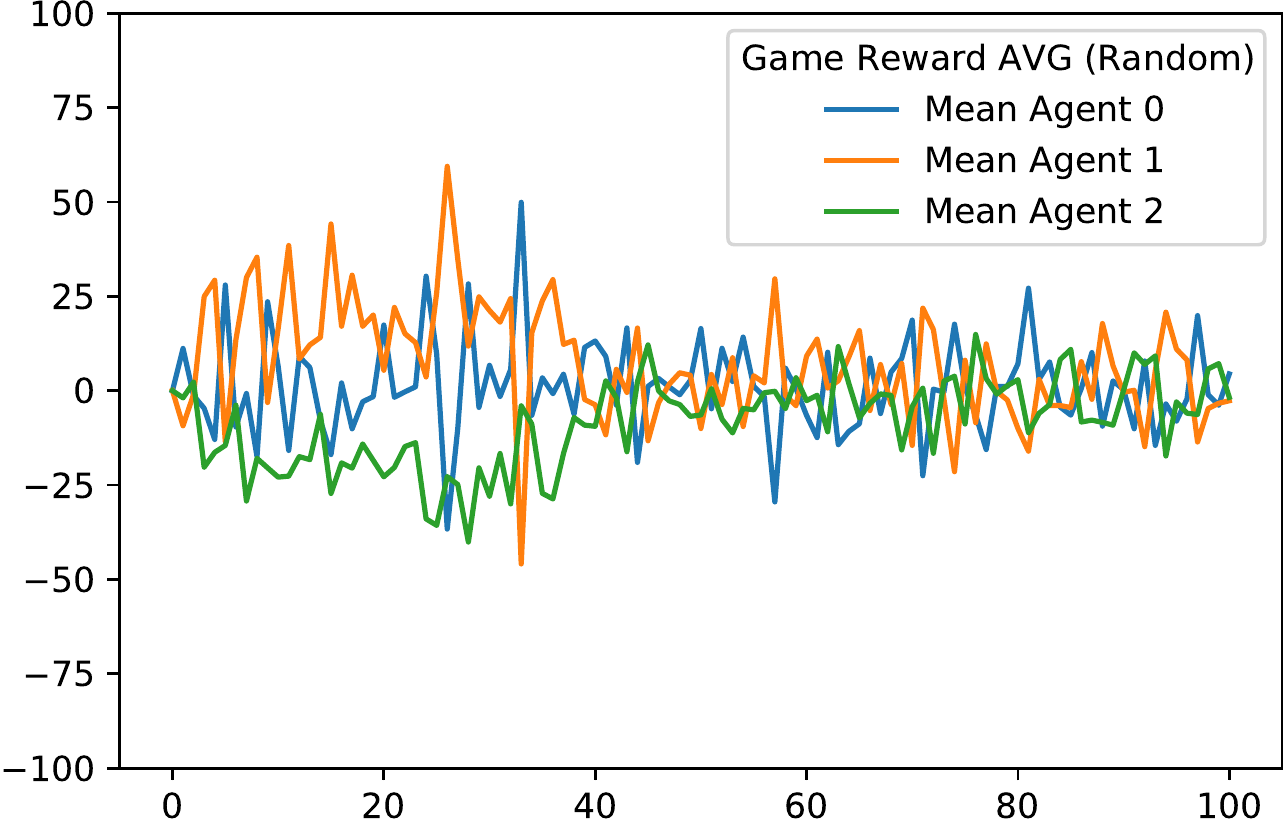}
    %\caption{Learning Rate 0.005}
  \end{minipage}
  \begin{minipage}[b]{0.49\textwidth}
    \includegraphics[width=\textwidth]{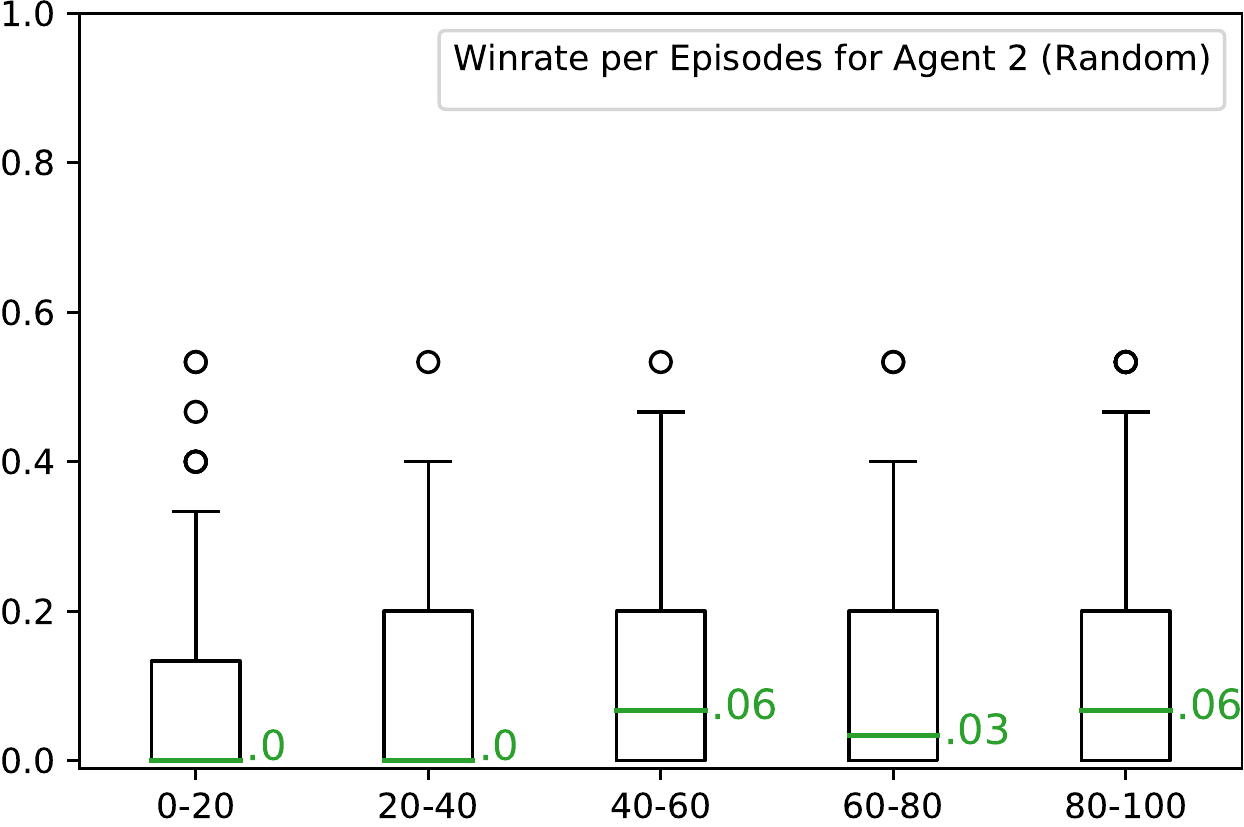}
    %\caption{Learning Rate 0.005}
  \end{minipage}
  \caption{Episode reward distribution within \textit{mode 2} including control runs where \(i_{f}\) is choosing random actions (lower half).} \label{collusionTwoAgentsImplicit}
\end{figure}

\vspace{-0.4cm}
\subsubsection{Mode 2: Implicit Communication}
The agents \(i_{c}\) successfully learned the suggested implicit collusion policy. After about 5 episodes, \(i_{c}\) achieve a higher game reward on average. This circumstance is especially prominent in the section between 20 episodes and 40 episodes (cf. the upper right half of figure \ref{collusionTwoAgentsImplicit}). On average, \(i_{f}\) is rarely able to exceed a reward of 0. Again, after about 40 episodes, the agents converge to an average game reward of 0.

We were able to observe a less prevailing, but still effective policy when implementing a randomly acting agent \(i_{f}\). As demonstrated in Figure 7, the median of \(i_{f}\)'s winrate was still 0 in between episodes 0 to 40, yet the interquartile range is greater than before, indicating a less stable learning due to the added randomness of \(i_{f}\)'s action selection. We also experience a few runs, where the agents were able to learn the displacement policy and not unlearn it in later episodes. In those specific runs, agent 1 repeated the same actions from episode 22 onward while agent 0 played the action that would lose against that one in a regular game. Hence, the joined rewards \(i_{c}\) turn out greater than those of \(i_{f}\).

\section{Discussion}
\vspace{\nudgeSections}
Our research successfully confirmed the hypothesis from law and economics scholars (e.\,g.,  \cite{Waltman.2008} or \cite{IZQUIERDO.2015}) about a possible collusion between reinforcement learning based pricing agents in MARL scenarios without being especially trained to do so. We furthermore extended these findings by providing specific learning stages that could be translated into real world scenarios to possibly set a foundation for a system that is capable to detect collusion in early stages. Moreover, we were able to show that with the appropriate reward function, deep reinforcement learning can be used to learn an effective displacement strategy in a MARL environment.

Based on the results of the experiments, we derive several implications. Due to the noticeable segmentation of action selection in different learning stages, one could argue that the transition episodes in between a fair and a collusive state can be seen as a signaling process, where agents agree on specific patterns to increase the joint reward. This proposition is supported by the fact, that a repeating action selection pattern of another agent could be predicted and punished by the DQN due to its experience replay \cite{L.Lin.1992}. In a real world scenario, a malicious AI could be trained to repeat patterns, that are less recognizable for humans. We would like to emphasize that within inelastic selling conditions (as they appear in collusive markets), a cooperation between two agents will be facilitated as the existing communication strategy will furthermore ease the displacement of a competitor. From a legal perspective, the question whether the cartel prohibition can be applied to such factual achieved, however non-volitional, state of collusion, is subject to this project's further legal research. 

\section{Limitations and Outlook}
\vspace{\nudgeSections}
As every study, the results are beset with limitations, opening the door for future research. As aforementioned, our experiment is a simplified, gamified version of an economy simulation game. As such, it lacks the data complexity of a real world pricing AI as well as the scaling opportunities. To further develop our research, we intend to apply the gained knowledge to a MARL environment resembling the one of real pricing AIs, where we can further highlight specific moments in which the agent's behavior tips over from independence to collusion. Especially the division into distinct stages should be investigated in a context of realistic pricing agents environments. While we focused on highlighting the possible dangers of pricing AIs in a MARL environment, we opened the opportunity for research explicitly investigating measures to avoid it. As such, law and IT scholars alike could benefit from this research as a foundation for guidelines, law amendments, or specific laws concerning the training and behavior of pricing AIs.   

\bibliographystyle{splncs04}
\bibliography{mybibliography}
\end{document}